# Convolutional Neural Network-based Topology Optimization (CNN-TO) By Estimating Sensitivity of Compliance from Material Distribution

Yusuke Takahashi
Tokyo Institute of Technology

Yoshiro Suzuki*
Tokyo Institute of Technology

Akira Todoroki
Tokyo Institute of Technology

**Abstract**

This paper proposes a new topology optimization method that applies a convolutional neural network (CNN), which is one deep learning technique for topology optimization problems. Using this method, we acquire a structure with a little higher performance that could not be obtained by the previous topology optimization method. In particular, in this paper, we solve a topology optimization problem aimed at maximizing stiffness with a mass constraint, which is a common type of topology optimization. In this paper, we first formulate the conventional topology optimization by the solid isotropic material with penalization method. Next, we formulate the topology optimization using CNN. Finally, we show the effectiveness of the proposed topology optimization method by solving a verification example, namely a topology optimization problem aimed at maximizing stiffness. In this research, as a result of solving the verification example for a small design area of $16 \times 32$ element, we obtain the solution different from the previous topology optimization method. This result suggests that stiffness information of structure can be extracted and analyzed for structural design by analyzing the density distribution using CNN like an image. This suggests that CNN technology can be utilized in the structural design and topology optimization.

## 1. Introduction

In recent years, deep learning (DL) has made it possible to recognize speech and images with high precision [1]. DL uses a deep neural network (DNN) in which a neural network (NN) is multilayered. In particular, the convolutional neural network (CNN), which is a type of DNN, has made great achievements in image recognition, and in December 2012, Le et al. succeeded in learning image features using CNN. In fact CNN showed that cats and human faces can be recognized [2]. In general, many methods are proposed for CNN to escape from the local solution of the error function so as to minimize the estimation error according to the task to be performed.

Structural optimization is a method for optimally designing structures under imposed conditions, or constraints, and there are three types. They are categorized as size optimization [3, 4], shape optimization [5, 6], and topology optimization (TO) [7, 8]. In TO, in addition to the outer shape of the structure, the topology of the structure, such as the number of holes, is also designed. Therefore, TO makes it possible to have a large degree of freedom in design variables compared to size optimization and shape optimization. By doing so, significant improvement is expected with regard to increasing performance requirements. However, the solid isotropic material with penalization (SIMP) method, which is a typical topology optimization method, has a problem in that an unambiguous solution cannot be obtained due to falling into the local optimal solution [7].

In this paper, we propose CNN-TO, which applies CNN to TO. In addition, because CNN has a large number of parameters and it is thought to fit well to the targeted task (high expressiveness). For that reason, CNN-TO is considered to be able to acquire a structure different from the structure acquired by the previous TO. Therefore, the purpose of this paper is to acquire the global optimum solution for topology by CNN-TO. In particular, in this paper, we deal with the structure optimization problem aimed at maximizing stiffness. This is because there are many cases where the stiffness maximization problem is applied to TO [9-13].

Next, we will refer to research similar to our study.

Nikola et al. proposed to implement 2D TO with a NN with a single fully-connected layer [14]. However, because the number of parameters for the target task in the single-layer NN is insufficient, the obtained TO result is not good.

A few existing studies address the use of CNN for TO.

Younggyu et al. constructed a CNN with an input of boundary condition and output of 2D TO result (optimized material density distribution) [15]. However, to construct the CNN, this method implements supervised learning as training data of the result of TO acquired by the SIMP method. That is, in this method, the same topology as the SIMP method is obtained as the optimization result.

Saurabh et al. showed that it is possible to obtain a solution similar to the SIMP method for the 3D TO problem by constructing a CNN with the input of boundary conditions and output of TO result [16]. Because this method also uses the results of TO obtained by the SIMP method as training data when constructing the CNN, the optimization result that can be obtained is also the same topology as the SIMP method.

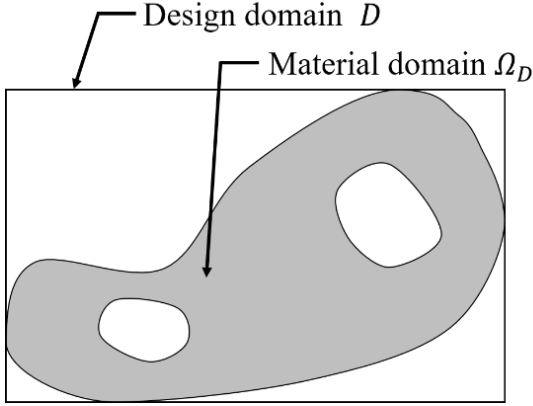

Fig. 1. Illustration of the solid design region $D$ and the region $\Omega_D$ distributing the material.

As mentioned above, the previous TO method using CNN imitated the TO solution of the SIMP method. In contrast, the TO method proposed in this paper searches for a solution different from the previous SIMP method by approximating the updating expression of the SIMP method by a CNN.

In this paper, we describe the previous TO in Section 2. Section 3 explains the TO using the proposed CNN (CNN-TO). Section 4 explains the example for performance verification handled in this paper. In Section 5, we compare the optimal solutions obtained by the previous TO and the new CNN-TO.

## 2. Previous topology optimization
## 2.1 Definition of topology optimization problem

In a TO problem, a solid design domain $D$ covering the domain $\Omega_D$ is set for distributing the material to form an optimum structure. The TO problem is a material distribution problem of arranging the domain $D \setminus \Omega_D$, which becomes a hole in the solid design domain $D$. The relationship between the solid design domain $D$ and the material domain $\Omega_D \Omega_D$ is illustrated as in Fig. 1.

In the material distribution problem, the Young's modulus $E$ is determined as shown in Eq. (2) using the characteristic function $\chi_\Omega$ defined by the following Eq. (1):

$$\chi_\Omega(\mathrm{x}) = \begin{cases} 1 & (\mathrm{x} \in \Omega_D) \\ 0 & (\mathrm{x} \in D \setminus \Omega_D) \end{cases} \quad (1)$$

$$E = E_0 \chi_\Omega \quad (2)$$

## 2.2 Analysis procedure of TO
## 2.2.1 Formulation of TO by SIMP method

In this paper, we propose a new CNN-TO and compare the effectiveness with the previous SIMP method. We will explain the previous method (general SIMP method) and CNN-TO in this section and subsequent sections.

The characteristic function $\chi_\Omega$ defined by Eq. (1) may have discontinuous points everywhere in the design domain. Homogenization methods [7] and the SIMP method are available for relaxing the discontinuity of $\chi_\Omega$. The SIMP method relaxes the discontinuity of $\chi_\Omega$ by introducing continuous density [17]. Compared with the homogenization method, the SIMP method is characterized by easy numerical calculation. In addition, because the SIMP method is applied to many optimization problems, it applies to the TO proposed in this research.

In the SIMP method, the solid design domain is modeled by the finite element method (FEM). After element division, a continuous density $x_i \in [x_{min}, 1]$ is assigned to each element, and the Young's modulus $E_i(x_i)$ of each element $i$ is calculated by the following Eq. (3).

$$E_i(x_i) = x_i^p E_0 \quad (3)$$

The lower limit of the density $x_{min}$ is set to eliminate numerical instability. The penalty parameter p in Eq. (3) is a value used to suppress $x_i$ from taking values other than 0 and 1 (intermediate density). The intermediate density and the method of suppressing it are described in detail in Section 2.3.

In the SIMP method, the stiffness optimization problem is formulated as the following Eq. (4):

$$\begin{aligned} &\text{find } \mathbf{x} = \{x_1, \cdots, x_N\} \\ &\min_{\mathbf{x}} c(\mathbf{x}) = \mathbf{u}^T \mathbf{f} \\ &\text{Subject} \\ &\mathbf{K}(\mathbf{x})\mathbf{u} = \mathbf{f} \\ &V(\mathbf{x}) = V_f \\ &x_{min} \leq x_i \leq 1, i = 1, \cdots, N \end{aligned} \quad (4)$$

## 2.2.2 Optimality criterion (OC) method

The problem expressed by Eq. (4) can be solved by the OC method. In the OC method, the continuous density $x_i$ is update by Eq. (5).

$$x_i^{new} = \begin{cases} \max(x_{min}, x_i - m) & \text{if } x_i B_i^\eta \leq \max(x_{min}, x_i - m) \\ \min(1, x_i + m) & \text{if } \min(1, x_i + m) \leq x_i B_i^\eta \\ x_i B_i^\eta & \text{else} \end{cases} \quad (5)$$

$B_i$ is a value determined by Eq. (6).

$$B_i = \frac{\partial c}{\partial x_i} / \Lambda \quad (6)$$

$\Lambda$ is a searched by a bisection method so that the total volume when updating the continuous density satisfies the mass constraint.

The algorithm of the OC method is to repeat the two operations of calculating sensitivity $\partial c / \partial x_i$ and updating $x_i$ according to Eq. (5) until the value of $x_i$ converges.

Although the OC method tends to obtain optimal solutions because sensitivity is used ($x_i$ tends to converge), it has been reported that there is a problem in that it is likely that the convergence will be a local optimal solution [1].

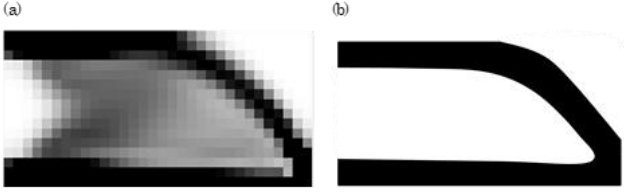

Fig. 2. Example of filtering in SIMP method: (a) Material density in which intermediate density exists. (b) Material density distribution when filtering excludes intermediate densities.

## 2.3 Intermediate density in SIMP method

A material having no density ($x_i$ of 0) indicates the absence of material, and an $x_i$ of 1 indicates the presence of material. In contrast, in SIMP method using the calculation procedure described in Section 2.2, the density $x_i$ can take values between 0 and 1 ($0 < x_i < 1$). This is called intermediate density. A structure that includes an intermediate density is difficult to actually manufacture. Therefore, various methods have been proposed to eliminate the intermediate density. Broadly speaking, these methods are classified into the following three types.

- Method of using spline curves to approximate representation of boundaries of designed members
  As a method of expressing boundaries using spline curves, a method of replacing TO with shape optimization problem using B spline curves was proposed by Yu et al. [18], Mingming et al. [19], and Yosef et al. [20].
- Method of using level set method without using the SIMP method
  There are several studies [21-24] of using the level set method without using SIMP method.
- Method using filtering
  Various filtering methods have been proposed by Sigmund [25].

In this paper, the SIMP method used for comparison with the proposed method uses the following filtering:

・Elimination of "floating islands" and
・Elimination of intermediate density.

The elimination of floating islands is described in section 3.2.4. For the elimination of the intermediate density, a threshold was set and simple filtering was applied (see Fig. 2), with material exsisting at densities above the threshold and voids existing at densities below the threshold. That is, when a design variable $x_i$ is input, a function that outputs the filtered variable $x_i^{\text{filtered}}$ after filtering is expressed by the following Eq. (7):

$$x_i^{\text{filtered}} = \begin{cases} 0, & x_i < L \\ 1, & x_i \geq L \end{cases} \quad (7)$$

Here, the threshold $L$ in Eq. (7) is set so that the total volume of the optimization result of the proposed method described in Section 3 is close to the total volume after filtering in the previous SIMP method.

## 2.4 Analysis procedure of SIMP method

As mentioned in Section 2.3, eliminating the intermediate density is meaningful for increasing the manufacturability of the structure. Therefore, the previous method for solving the example for performance verification in this paper is based on the following analysis procedure:

1. Generate the initial structure
2. Perform FEM analysis of current structure (density $x_i$)
3. Calculate the sensitivity of compliance to the current structure (density $x_i$)
4. Update the density based on the OC method expressed by Eq. (5) to obtain $x_i^{new}$
5. Repeat steps 2 to 4 until the difference between $x_i^{new}$ and $x_i$ falls below a certain value to obtain the optimal solution
6. Apply the filtering expressed by Eq. (7) to the optimal solution

Even if the optimum solution obtained by the OC method is a structure including some intermediate density that is difficult to manufacture, applying the filter eliminates the intermediate density and produces a structure that can be manufactured.

## 3 Topology optimization using CNN (CNN-TO)
## 3.1 Overview of CNN

CNN is a NN using a convolution layer that reduces the degree of freedom of the connection weight as compared with the fully connected NN [1]. The idea of CNN comes from neocognitron [26] and LeNet [27], and it existed since the 1990s. The CNN proposed based on those ideas has recently been attracting attention [1] as a result of having shown a major difference from other methods in a general object recognition competition in 2012 [28].

The calculation of convolution in flat-convolution (strides = 1 convolution) is shown in Eq. (8).

$$y_{u,v} = b_{u,v} + \sum_{i=-k'_h}^{k'_h} \sum_{j=-k'_w}^{k'_w} W_{i+k'_h, j+k'_w} x_{u+i, v+j} \quad (8)$$

where $k'_h, k'_w$ are $k'_h = (k_h - 1)/2$, $k'_w = (k_w - 1)/2$. In the convolution layer, $W_{i,j}$ becomes a parameter, which is appropriately determined in the process of learning. $b_{u,v}$ is also decided appropriately in the learning process.

For down-convolution with stride set to 2 or more, up-convolution is set to less than 1

## 3.2 Analysis procedure of proposed method

The procedure of CNN-TO proposed in this paper is as follows:

1. Acquire training data
2. Build the CNN
3. Generate the initial structure
4. Calculate sensitivity
5. Update the OC method expressed by Eq. (5) until convergence
6. Apply filtering

CNN-TO constructs a CNN that uses material density distribution as input and sensitivity as output. That is, the CNN is used for sensitivity calculation in the SIMP method, which is different from the previous SIMP method. By using CNN for sensitivity calculation, it is expected that an optimal solution different from the previous method can be obtained.

Each step will be described in detail below.

### 3.2.1 Acquisition of training data

The type of training data used has a very important effect on the accuracy of CNN. When a material density distribution is created using random numbers and used as learning data, a structure without connectivity, as shown in Fig. 3, is included in the training data, and a structure difficult to manufacture is also learned. Learning structures that are difficult to manufacture seems to be a hindrance in acquiring a manufacturable structure. Therefore, when preparing learning data, it is necessary to generate many structures with continuity using random numbers.

Techniques for expressing a structure with continuity are described below.

Wang et al. proposed a bar-system representation method [29] to express the material density distribution as bars (see Fig. 4) based on graph theory. the bar-system representation method generates multiple bars to connect the support point and the load point. Then, for each generated bar, the width and the coordinates of the end point are given as design variables. When designing the design domain of TO by FEM, if the center of gravity position of each element is included in the area where the bar exists, the density 1 is given to that element ($x_i = 1$).

In the bar-system representation method, as shown in Fig. 5, because the presence or absence of a material is expressed in a matrix form, compatibility is good when inputting it to a CNN, which is good at image recognition. For this reason, we use the bar-system representation method in this study.

To make CNN supervised learning, it is necessary to prepare sets of material density distributions and sensitivities. Sensitivity is a partial differentiation of compliance as an objective function with density, which is a design variable. Sensitivity is computed by FEM analysis.

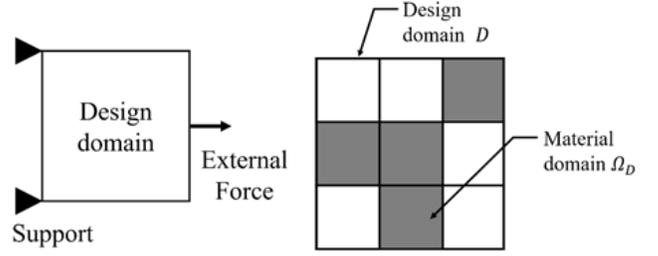

Fig. 3. Improperly connected structures in the bit-array representation method.

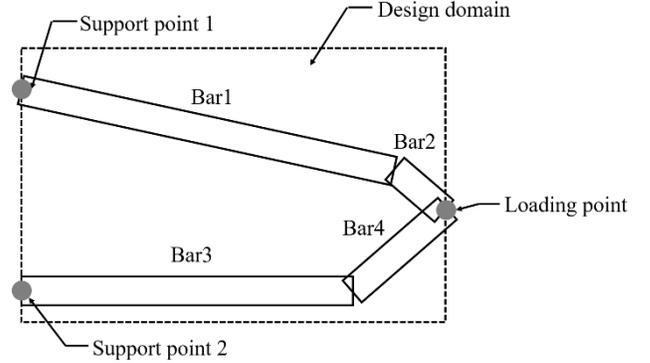

Fig. 4. Illustration of bar-system representation.

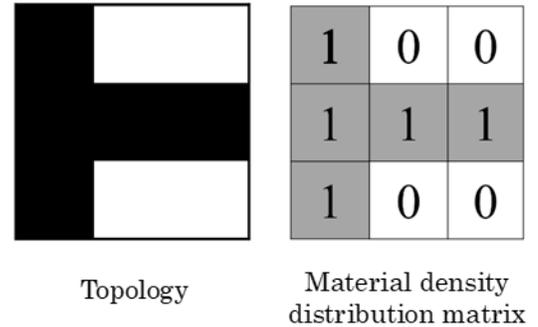

Fig. 5. Result of bar-system representation.

### 3.2.2 Construction of CNN

The CNN is trained by supervised learning using the training data mentioned in section 3.2.1. CNN input is material density vector **x**, and its output is sensitivity $\partial c / \partial \mathbf{x}$ of compliance $c$. That is, the CNN estimates a function $f$ that satisfies the following Eq. (9):

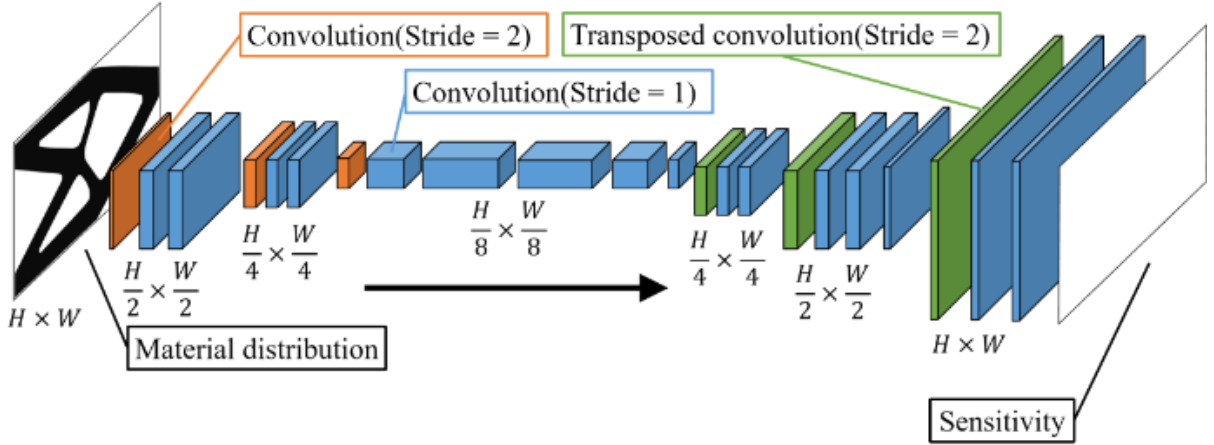

Fig. 6. Overview of proposed CNN model.

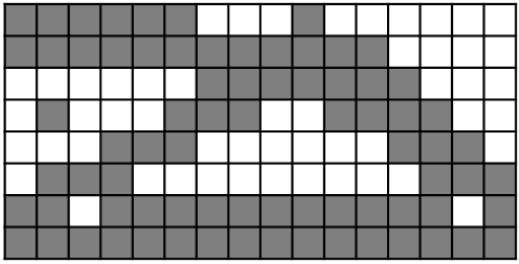

Fig. 7. Optimal structure generated by CNN. The presence of locally discontinuous elements mean that the overall structure is not realistic.

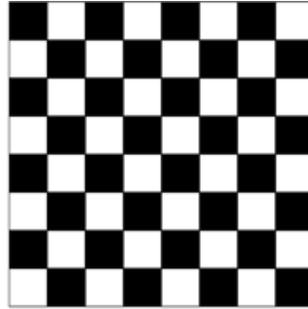 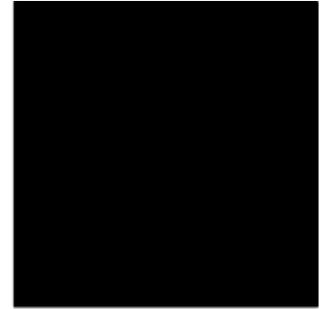

(a) Design variables before filtering (checkerboard pattern has occurred).

(b) Design variable after filtering (although the checkerboard pattern has been eliminated, the volume has changed greatly).

Fig. 8. (a) Example of checkerboard pattern. (b) Example after filtering. Checkerboard pattern is excluded by carrying out filtering in the proposed method.

$$\frac{\partial c}{\partial \mathbf{x}} = f(\mathbf{x}) \qquad (9)$$

The composition of the constructed CNN is shown in Table 1 and Fig. 6. This CNN is mainly composed of three types of convolution layers: flat-convolution, up-convolution, and down-convolution. CNNs using down-convolution and up-convolution are used in fields such as rough sketch line drawing [30] and deep convolutional generative adversarial networks [31].

Simoserra et al. used down-convolution and up-convolution to construct a CNN that achieved the rough sketch line drawing task and included the influence of a small-scale structure (fine lines) in the sketch. This is reported to help achieve beautiful line drawing [30].

In this research, the use of down-convolution and up-convolution plays the role of suppressing the occurrence of small-scale structures with low manufacturability in the optimization result. This is thought that it contributes to outputting a simple optimization result that is easy to manufacture.

### 3.2.3 Optimum solution obtained by OC method

The constructed CNN plays the role of function $f$ in Eq. (9). The material density distribution of Eq. (5) is updated using the approximate value of the sensitivity calculated using the constructed CNN, thereby optimizing the material density.

In general, however, the estimation accuracy of CNN is not 100%, and the estimation includes errors. Although this error causes the optimal structure generated by CNN to have an excellent shape as an overall outline, there are many locally discontinuous parts; in other words, the result might not be a mechanically meaningful and realistic structure (see Fig. 7). Therefore, to acquire an optimum structure that is meaningful from the viewpoint of engineering, it was necessary to enhance the continuity of the structure by applying the filtering described in the next section with the optimum structure generated by the CNN as the initial structure. The applied filtering algorithm is described in section 3.2.4.

### 3.2.4 Filtering in the proposed method

Table 1 Architecture of CNN used in this paper

| Layer number | Type | Stride | Output size |
|---|---|---|---|
| 1 | Input | - | $1 \times H \times W$ |
| 2 | Down-convolution | $2 \times 2$ | $48 \times H/2 \times W/2$ |
| 3 | Flat-convolution | $1 \times 1$ | $128 \times H/2 \times W/2$ |
| 4 | Down-convolution | $2 \times 2$ | $256 \times H/4 \times W/4$ |
| 5 | Flat-convolution | $1 \times 1$ | $256 \times H/4 \times W/4$ |
| 6 | Flat-convolution | $1 \times 1$ | $256 \times H/4 \times W/4$ |
| 7 | Down-convolution | $2 \times 2$ | $512 \times H/8 \times W/8$ |
| 8 | Flat-convolution | $1 \times 1$ | $512 \times H/8 \times W/8$ |
| 9 | Flat-convolution | $1 \times 1$ | $1024 \times H/8 \times W/8$ |
| 10 | Flat-convolution | $1 \times 1$ | $1024 \times H/8 \times W/8$ |
| 11 | Flat-convolution | $1 \times 1$ | $1024 \times H/8 \times W/8$ |
| 12 | Flat-convolution | $1 \times 1$ | $1024 \times H/8 \times W/8$ |
| 13 | Flat-convolution | $1 \times 1$ | $512 \times H/8 \times W/8$ |
| 14 | Up-convolution | $2 \times 2$ | $512 \times H/4 \times W/4$ |
| 15 | Flat-convolution | $1 \times 1$ | $256 \times H/4 \times W/4$ |
| 16 | Flat-convolution | $1 \times 1$ | $256 \times H/4 \times W/4$ |
| 17 | Up-convolution | $2 \times 2$ | $256 \times H/2 \times W/2$ |
| 18 | Flat-convolution | $1 \times 1$ | $128 \times H/2 \times W/2$ |
| 19 | Up-convolution | $2 \times 2$ | $48 \times H \times W$ |
| 20 | Fully connected layer | | |

For filtering in the proposed method, we considered the following procedure:

・Elimination of floating islands
・Filling of small holes
・Elimination of "peninsulas"

The TO might produce an optimum structure that has a checkerboard pattern or other excessively complicated structure [30]. The checkerboard pattern is a phenomenon resulting from inappropriateness of numerical analysis, which is caused by overestimating the rigidity in the shear direction (known as shear locking phenomenon) in the FEM. In addition, the structure including a checkerboard pattern contains excessively complicated partial local structures and it is difficult to manufacture (see Fig. 7). Therefore, it is necessary to remove these patterns properly.

The filtering described in this section fills minute holes in the structure to suppress checkerboard pattern generation (see Fig. 8).

However, when the checkerboard pattern is regarded as a small hole and filled, this action change the mass of material density in the optimization result. As a result, it is assumed that implementation of filtering will not satisfy mass constraints set in TO. Therefore, when implementing the filtering described in this section, it should be checked whether a checkerboard pattern is included in the optimization result.

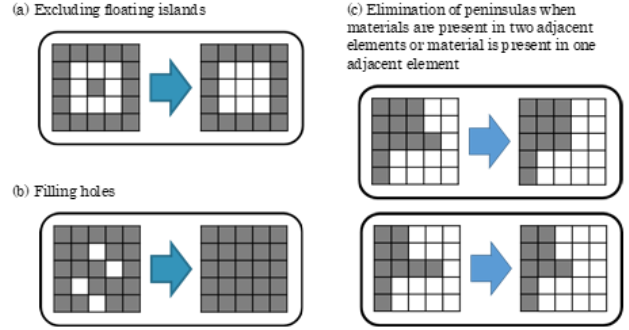

Fig. 9. Conceptual diagram of filtering applied in the proposed method.

#### 3.2.4.1 Elimination of floating islands

In the optimum structure generated by CNN, floating islands, which are not connected to any other elements regarded as material, may occur. These floating islands are impossible to manufacture and do not contribute to the improvement of rigidity at all, so they must be eliminated (see Fig. 9 (a)).

#### 3.2.4.2 Filling of small holes

In the optimum structure generated by CNN, small holes may be generated. Small holes are difficult to manufacture and should be eliminated (see Fig. 9 (b)). However, holes with sizes larger than a certain size are not excluded. The reason for this is that large holes do not necessarily reduce manufacturability and may contribute to improvement in rigidity.

#### 3.2.4.3 Elimination of peninsulas

In the optimum structure generated by CNN, peninsulas may occur, in which the number of elements in which the material exists is 2 or less among 8 adjacent elements. Peninsulas are structures with low contribution to rigidity improvement, and they should be eliminated (see Fig. 9 (c)).

### 4. Example for performance verification

In this section, we describe an example TO to verify the performance of the proposed method. A design domain of 16 × 32 square elements is targeted. As the boundary condition, we choose the cantilever. We deal with the stiffness maximization problem represented by Eq. (4). Also note again that the stiffness maximization problem is synonymous with the compliance minimization problem. In the proposed method described in Section 3 and the previous SIMP method described in Section 2.4, material densities that maximize rigidity are acquired and the results are compared.

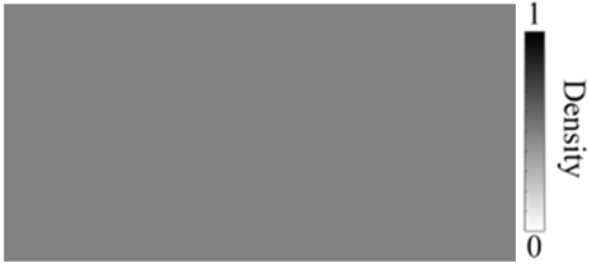

Fig. 10. Initial material density distribution used for optimization. This was used for both the previous method and proposed method.

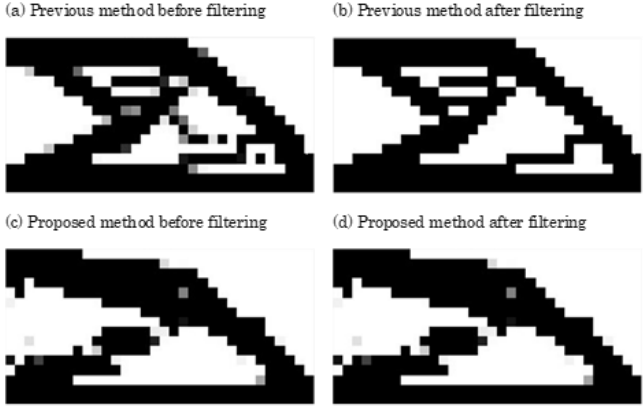

Fig. 11. Material density distribution that maximizes rigidity. (a) Material density distribution obtained before filtering with previous method. (b) Material density distribution after filtering with previous method. (c) Material density distribution before filtering with proposed method. (d) Material density distribution after filtering with proposed method. For all, results are shown for 40 iterations.

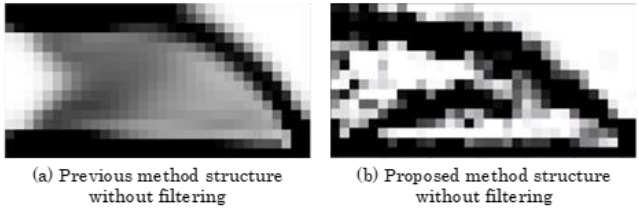

Fig. 12. Material density distribution after five updates. (a) Previous method structure after five updates (raw data acquired by optimality criteria method without filtering) (b) Proposed method structure after five updates (raw data acquired by optimality criteria method without filtering). For all, results are shown for 5 iterations.

Upon optimization, the upper limit of the volume is set as 50% with respect to the volume of the entire design domain, and the penalty parameter $p$ in Eq. (3) was set to 3. In addition, the initial material distribution during optimization was assumed to have an evenly distributed density over the design area (see Fig. 10).

Table 2 Comparison of calculation time of previous method and proposed method

| Method | Calculation time (s) | Density update count | Computing environment |
|---|---|---|---|
| Previous method | 7.28 | 40 | CPU |
| Proposed method | 33.5 | 40 | CPU+GPU |

Table 3 Comparison of compliance of initial design, optimal design obtained by the previous method (SIMP), and optimal design by the proposed method.

| | Initial design | Optimal design obtained by SIMP method | Optimal design obtained by proposed method |
|---|---|---|---|
| Dimensionless compliance | 5,974 | 1,294 | 1,287 |

In the FEM analysis, Young's modulus $E_0 = 1.0$, Poisson's ratio $\nu = 1.0$, and density $\rho = 1.0$ in the region where the material is arranged ($x_i = 1$). In the rigidity maximization problem, because the relative value of how much the compliance has changed from the initial structure by optimization is important, a simple physical property value was set and the unit was omitted.

For the previous method, a PC equipped with 8 GB RAM and Core i5-4500 U @ 1.60 GHz CPU was used for the calculation without parallel calculation by GPU. For the proposed method, we used a 128-GB PC with Xeon Silver 4112 @ 2.60 GHz CPU for the calculation. In the proposed method, the CNN is calculated by parallel computation with a GPU (NVIDIA GeForce GTX 1080 Ti × 4).

The Python programming language was used with the library Tensorflow for CNN calculation and NumPy and SciPy for other calculations.

In this research, we use Adam as an optimizer. Also, Adam's hyperparameters were set to learning_rate = 0.00005, beta1 = 0.9, beta2 = 0.999, and epsilon = 1e-8. For details on each hyper parameter, see [32].

## 5. Analysis result and consideration
### 5.1 Analysis result

The results of solving the stiffness maximization problem described in Section 4 by the previous SIMP method are shown in Fig. 11 (a) and (b), and the results produced by the proposed method are shown in Fig. 11(c) and (d). Fig. 11(a) and (c) shows the material density distributions before filtering, and Fig. 11(b) and (d) shows the material density distributions after filtering.

### 5.2 Comparison between proposed method and previous method
### 5.2.1 Comparison in calculation time

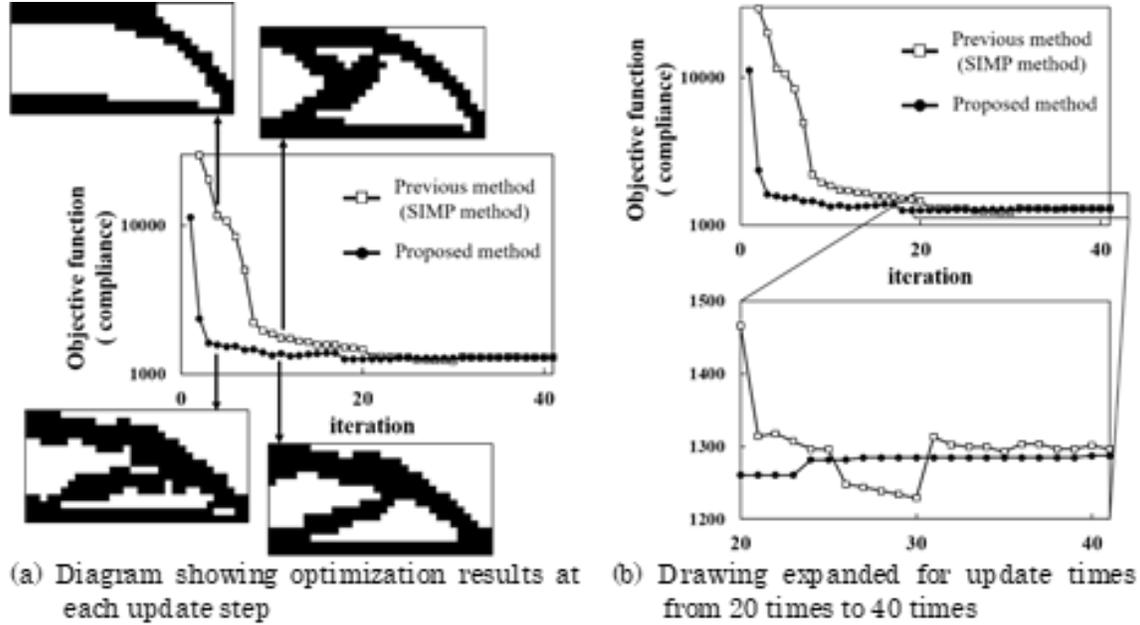

(a) Diagram showing optimization results at each update step

(b) Drawing expanded for update times from 20 times to 40 times

Fig. 13. Details of optimization progress at each step. The horizontal axis is the number of updates of the material density distribution. The vertical axis is the value of the objective function.

Table 2 shows the comparison of the computation time required for implementing the same number of density updates (see Eq. (5)) for the previous method and the proposed method.

As shown in Table 2, it was found that the proposed method requires more computation time than the previous method. This is thought to be caused by the fact that the CNN constructed this time is multilayered and requires time for inference.

However, compared with the previous method, it was found that the proposed method had few update times required to obtain the outline of the structure. Fig. 12(a) shows the structure density after five updates, when the initial structure has a uniformly distributed density throughout the design domain (see Fig. 10); the same results for the proposed method are shown in Fig. 12(b). In comparison with the optimization result obtained by the previous method (see Fig. 12), the optimization result (see Fig. 12 (b)) obtained by the proposed method obtains different results even with the same number of updates. In the proposed method, the outline of the structure has already been acquired after five updates. In contrast, in the previous method, there are many elements of gray, and the outline of the structure has not yet emerged.

Fig. 13 shows the previous method and the proposed method, with the number of updates plotted on the horizontal axis and the value of the compliance on the vertical axis. As shown in Fig. 13, the proposed method reduces the compliance with a smaller number of updates compared with the previous method. In Fig. 13(b), there is a portion where the value of compliance increases as the number of updates increases (corresponding to about 30 updates in the previous method). This is due to the fact that the filtering used in the proposed method and the previous method does not necessarily have only the effect of decreasing the compliance.

### 5.2.2 Comparison of optimal solution performance

Table 3 shows the compliance of the optimal solutions (Fig. 11(b) and Fig. 11(d)) obtained for the previous method and the proposed method. Table 3 also shows the compliance of the initial solution at the time of optimization.

The performance (compliance) for the optimal solutions obtained by the previous method and proposed method was almost equal. However, the obtained topology was different.

The optimal solution obtained by the proposed method has a simple structure compared with the optimal solution obtained by the previous method. Even if the initial solution was given randomly and the optimization by the previous method was executed more than once, a structure similar to the optimum solution of the proposed method could not be acquired.

From the above, it can be said that the optimal solution obtained by the proposed method was a structure that cannot be acquired by the previous method.

### 5.3 Consideration of analysis results
### 5.3.1 Reasons for obtaining optimal solutions different from previous methods by the proposed method

In this section, we discuss the reason why the proposed method could acquire a structure that could not be acquired by the previous method as the optimal solution.

The OC method, which is the optimization method used in this paper, is a method to solve the optimization problem by using the gradient of the objective function. Therefore, the OC method is modeled as a gradient.

When minimizing the objective function using the gradient method, that is, when solving the optimization problem expressed by the following Eq. (10), the update equation of the gradient method is expressed by Eq. (11):

$$\min_{\mathbf{x}} f(\mathbf{x}) \qquad (10)$$

$$\mathbf{x}^{(t+1)} = \mathbf{x}^{(t+1)} - \varepsilon \frac{\partial f}{\partial \mathbf{x}^{(t)}} \qquad (11)$$

where $f(\mathbf{x})$ is an objective function aiming at minimization and $\mathbf{x}$ is a vector having a design variable as its component.

In the update formula (Eq. (5)) of the proposed method in this research, we use the approximate value obtained by CNN for the gradient of the objective function. Generally, estimation of CNN includes errors. Therefore, the updating expression in the proposed method can be regarded as an expression of the gradient method that include a random component in the gradient.

Generally, it is known that by including a random component in the updating expression in the gradient method, it is possible to perform optimization that is unlikely to fall into a local solution. The update formula in the proposed method also includes a random component in the update formula, so optimization can be performed without falling into a local solution. As a result, it is considered that an optimum solution different from that of the previous method could be obtained. In our verification example, the error in estimation using CNN worked well. However, it is necessary to consider in the future whether the same results as this study can be obtained for any verification example.

### 5.3.2 Stability of the proposed method

In this section, we investigate to what extent the structure obtainable by the proposed method can be reproduced.

Generally, it is known that the performance of CNN has low reproducibility [33]. In this study, because structure optimization is carried out by estimating the sensitivity with CNN, it is considered that the structure that can obtain the performance of CNN is influenced, and the investigation of reproducibility is important.

Regarding the performance of CNN, the reason for poor reproducibility is that the number of parameters is enormous in multilayered CNNs, and the weighting applied during CNN construction contains many local optimal solutions [1]. Because there are many local optimal solutions, even if similar hyperparameters are set, the possibility of falling into a different local optimum solution during learning is high, and the reproducibility is low. Various methods have been proposed to enhance the performance of CNN. Some examples are:

- Because the performance of CNN is known to depend on the initial weighting (initial parameter values) [34], there are methods for appropriately setting the initial weight [35,36]
- Optimization method for optimizing CNN parameters [32,37-40]
- Dropout [41]
- Batch normalization [42]

In this research, we use Adam as the optimizer because the best result was obtained when changing the optimization method.

The CNN having the structure shown in Table 1 was taught twice with Adam using the hyperparameters described above while changing the initial weight. For convenience, one is called CNN 1 and the other is called CNN 2. Fig. 14(a) shows the optimal solution obtained by the proposed method using CNN 1, and Fig. 14(b) shows the optimal solution obtained by the proposed method using CNN 2. It can be seen that there is similarity between the optimization results acquired by CNN 1 and CNN 2. However, because the same optimum solution was not obtained, the CNN performance reproducibility is not necessarily high. From this, it can be said that the stability of the proposed method has room for improvement.

### 5.3.3 Advantage of the proposed method

In this section, we describe the superiority of the proposed method compared to the previous method. The advantages of the proposed method are as follows:

・**Acquire optimal solution with fewer updates**

As described in Section 5.2.1, compared to the previous method, the proposed method can acquire the optimal solution with a small number of updates. When the design area becomes large, it is considered that it takes a long time to calculate the sensitivity in the optimum solution search. At that time, it is an advantage to be able to acquire an optimal solution with a small number of updates.

・**Design a simpler structure**

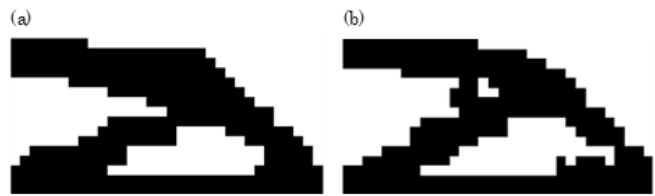

Fig. 14. (a) Optimization result obtained by proposed method using CNN 1 (b) Optimization result obtained by proposed method using CNN 2.

In the proposed method, a hole can be created in the solid design domain during the search for the optimal solution, so it is classified as TO. TO is performed as the conceptual design stage [43]. Therefore, as described in Section 5.2.2, using the proposed method to acquire a structure that cannot be acquired by the previous method as the optimal solution provides a new tool for the designer who performs the conceptual design.

## 6. Conclusion

We proposed a topology optimization method using CNN (CNN-TO) and attempted to acquire a structure that could not be acquired with the previous topology optimization method (previous SIMP method).

To confirm the effectiveness of the proposed method, we used a rigidity maximization problem in a rectangular design domain. In our study, a cantilever beam was selected as the boundary of the verification example.

As a result, we clarified that the proposed method can acquire the optimal solution that could not be obtained by the previous SIMP method, and that the optimal solution can be obtained with fewer update steps than the previous SIMP method. Obtaining a solution different from that of the previous SIMP method was impossible when using a conventional fully connected NN for topology optimization. Therefore, the significance of using CNN for topology optimization was shown.

However, CNN has a problem in that its performance is not stable due to falling into local solutions of the error function during network construction. Due to this problem, it was shown that the same optimum solution cannot necessarily be acquired again, even when the same CNN architecture hyperparameters are set.

In future, it is thought that it is necessary to improve the stability during CNN construction by implementing optimization with the dropout regularization technique, optimizers, and different hyperparameter values. In our verification example, the error in estimation using CNN worked well.

However, it is necessary to consider in the future whether the same results as this study can be obtained for any verification example.

## Acknowledgments

This work has been supported by the Tokyo Tech Assistant Professor Incentive Award 2018.

## Appendix
### Appendix 1. Validity of using different filters in previous and proposed methods

As described in Sections 2.3 and 3.2.4, different filters were applied to both the previous and proposed methods.

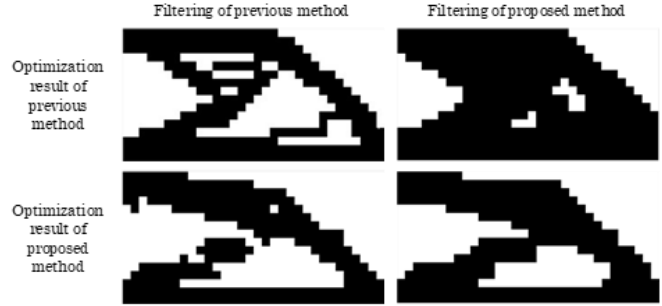

Fig. 15. Effects of changing filtering type on optimization results.

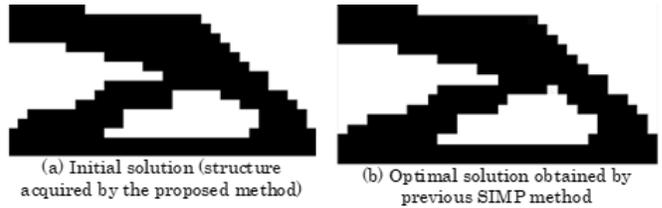

Fig. 16. Results of using proposed method results for previous method input. The initial solution is the structure acquired by the proposed method, and the optimal solution is that obtained when the previous method is applied to it.

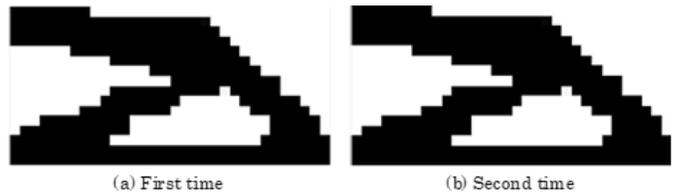

Fig. 17. Optimal solutions acquired in multiple runs with the same CNN.

This is because when the filtering used in the proposed method is applied to the optimization result obtained by the previous method, the area of the structure remarkably increases and it is difficult to compare with the optimization result obtained by the proposed method.

When applying the filtering used in the previous method to the optimization result acquired by the proposed method, a structure with low rigidity was obtained because the optimization result produced a structure with low continuity.

For the above reasons, different filters was used for the previous and proposed methods. Fig. 15 shows the results of applying these filters to the optimization results obtained by the previous and proposed methods.

### Appendix 2. Using the optimization result of the proposed method as the initial solution of the SIMP method

In the previous SIMP method, TO is performed using the gradient of the objective function calculated by FEM analysis. By adopting the optimal solution acquired by the proposed method as the initial solution for implementation in the previous method, we can expect to acquire a structure with further improved performance, while maintaining similarity with the optimal solution obtained by the proposed method. Fig. 16(b) shows the result of initiating the previous method with the optimal solution (shown in Fig. 16(a)) obtained by the proposed method. By comparing Fig. 16(a) and (b), we can see that the structure has not changed substantially. In other words, the optimal solution obtained by using the proposed method that performs approximate gradient calculation using CNN is one of the local optimal solutions that can be reached by the previous method (there is a possibility of a global optimal solution).

## Appendix 3. Reproducibility of the proposed method

Figure 16 shows the results when multiple optimizations are performed for the case where the same CNN is used. Because the structures in Fig. 17(a) and (b) are identical, it was confirmed that the optimal solution that can be obtained does not change even when the proposed method is optimized multiple times.

**Nomenclature**

| | |
|---|---|
| $b_{u,v}$ | A value called bias |
| $C$ | Index value of rigidity called average compliance |
| **f** | Nodal load vector |
| **K** | Stiffness matrix |
| $k_h$ | Height of the convolution filter |
| $k_w$ | Width of the convolution filter |
| $m$ | Move limit |
| **u** | Nodal displacement vector |
| $V$ | Total volume |
| $Vf$ | Upper limit of the total volume |
| $W_{i,j}$ | Pixel value of the convolution filter |
| **x** | Vector in which the material density $x_i$ of each element in the design domain is arranged |
| $x_{u,v}$ | Input to the convolution layer |
| $y_{u,v}$ | Value output by the convolution layer |
| $\eta$ | Damping coefficient |
| $\Lambda$ | Undetermined Lagrange multiplier |